\documentclass{esannV2}
\usepackage{graphicx}
\usepackage{subfig}
\usepackage{url} 
\usepackage[latin1]{inputenc}
\usepackage{amssymb,amsmath,array}
\usepackage{hyperref} 
\usepackage[table]{xcolor}
\usepackage[export]{adjustbox}

\begin{document}
\title{Data-driven Verification of DNNs for Object Recognition}

\author{Clemens Otte$^1$, Yinchong Yang$^1$, and Danny Benlin Oswan$^{1,2}$  
\vspace{.3cm}\\
1- Siemens AG - Technology, 81739 Munich, Germany
\vspace{.1cm}\\
2- Technical University of Munich, Informatics, 80333 Munich, Germany\\
}

\maketitle

\begin{abstract}
The paper proposes a new testing approach for Deep Neural Networks (DNN) using gradient-free optimization to find perturbation chains that successfully falsify the tested DNN, going beyond existing grid-based or combinatorial testing. Applying it to an image segmentation task of detecting railway tracks in images, we demonstrate that the approach can successfully identify weaknesses of the tested DNN regarding particular combinations of common perturbations (e.g., rain, fog, blur, noise) on specific clusters of test images.
\end{abstract}

\section{Introduction}
Deploying Deep Neural Networks (DNN) in safety-critical fields such as autonomous vehicles \cite{fingscheidt_deep_2022}
requires verifying their robustness and reliability. Formal methods \cite{liu_algorithms_2021, meng_adversarial_2024} 
are usually not applicable to the large models used in real-world applications.  Thus, verifying the DNN's robustness is mostly done with systematic empirical testing. For example, \cite{chand2021_combinatorial} uses combinatorial (pairwise) testing, which, however, leaves many configurations untested. 

Adversarial robustness \cite{meng_adversarial_2024} considers maliciously contrived imperceptible input perturbations, which are often created by following the gradient of the model error on the input pixels.
Usually, the likelihood of encountering such adversarial inputs in reality is low and a more practical concern is the robustness towards common (natural) perturbations \cite{hendrycks2019robustness}, e.g., blur, fog, snow, rain, noise, and geometric distortions. The question we address in the following is whether these common perturbations can be combined in an adversarial manner that falsifies the tested DNN. Since natural perturbations are in general not differentiable, gradient-based optimization is not applicable. Instead, we propose a new testing approach utilizing gradient-free optimization to guide 
the perturbation parameters to regions where the model output shows 
larger model errors, thereby increasing the likelihood of generating test 
cases that falsify the model. The model output is used as feedback to the optimizer, bypassing
the need for gradient information or the DNN's architecture, making the
approach scalable for use on large contemporary DNNs. 

\begin{figure}%
    \centering
    \subfloat[\centering Original]{{\includegraphics[width=.23\textwidth]{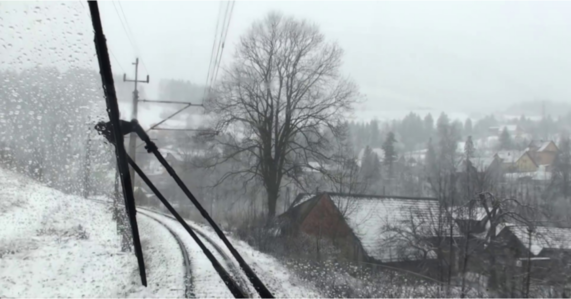} }\label{fig:ex1_orig}}%
    \hfill
    \subfloat[\centering IoU = 0.95]{{\includegraphics[width=.23\textwidth]{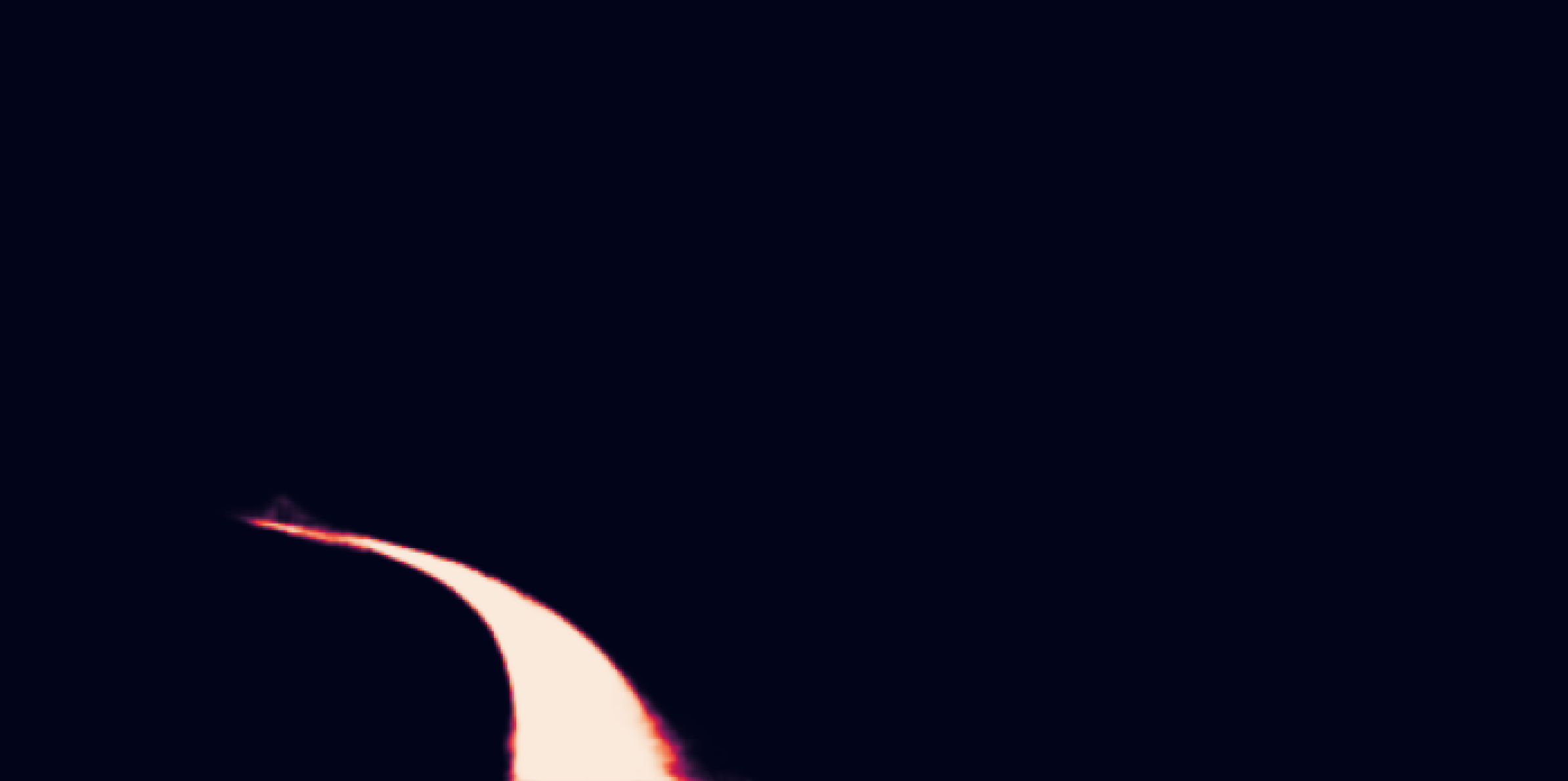} }}%
    \hfill
    \subfloat[\centering Perturbed]{{\includegraphics[width=.23\textwidth]{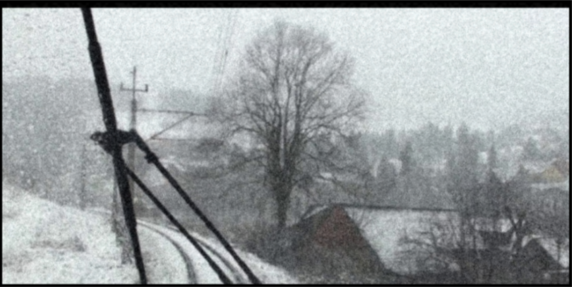} }}%
    \hfill
    \subfloat[\centering IoU = 0]{{\includegraphics[width=.23\textwidth]{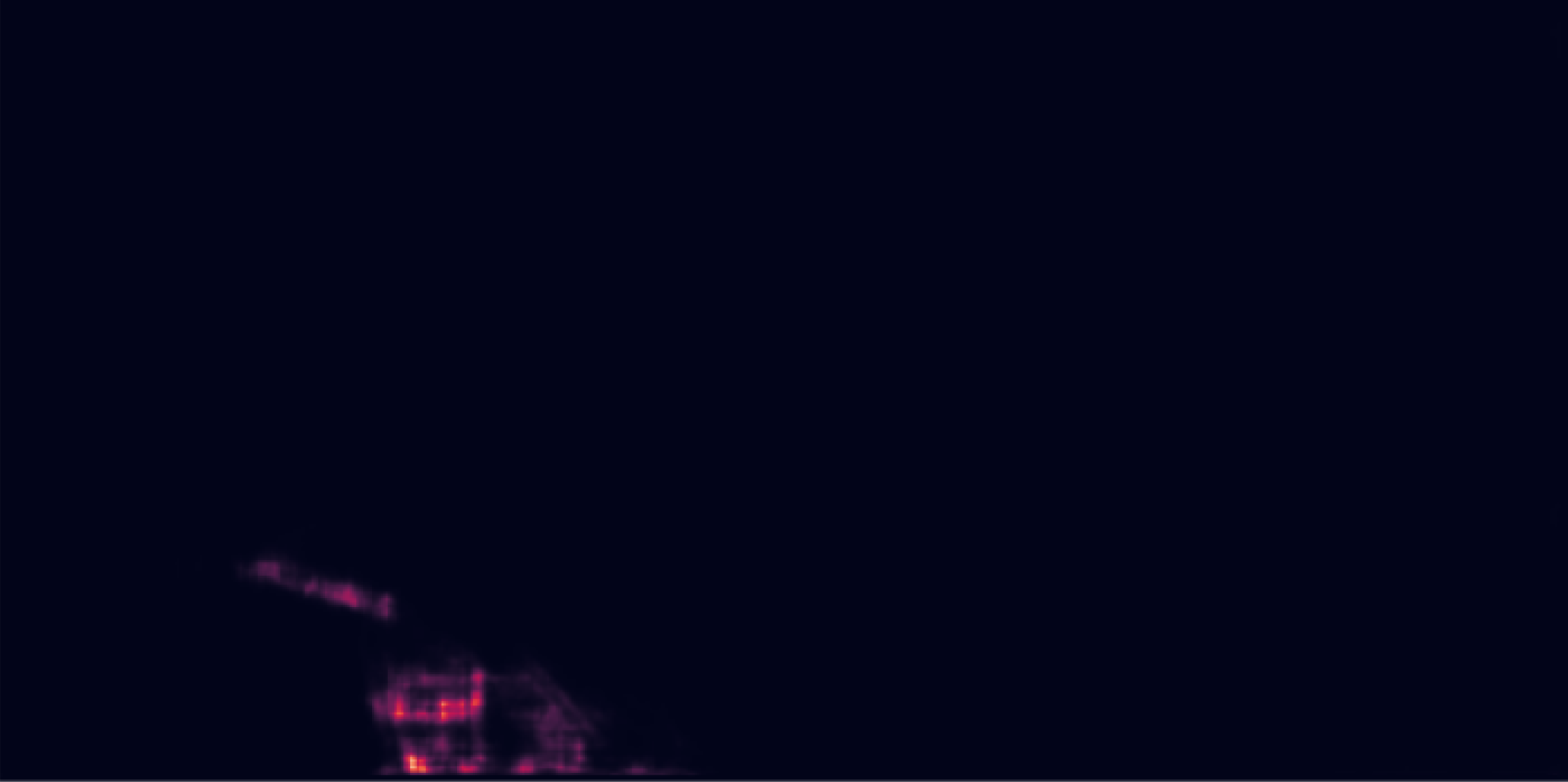} }}\\[-3mm]
    \subfloat[\centering Original]{{\includegraphics[width=.23\textwidth]{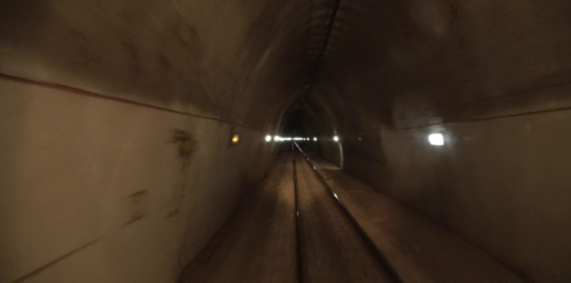} }}%
    \hfill
    \subfloat[\centering IoU = 0.98]{{\includegraphics[width=.23\textwidth]{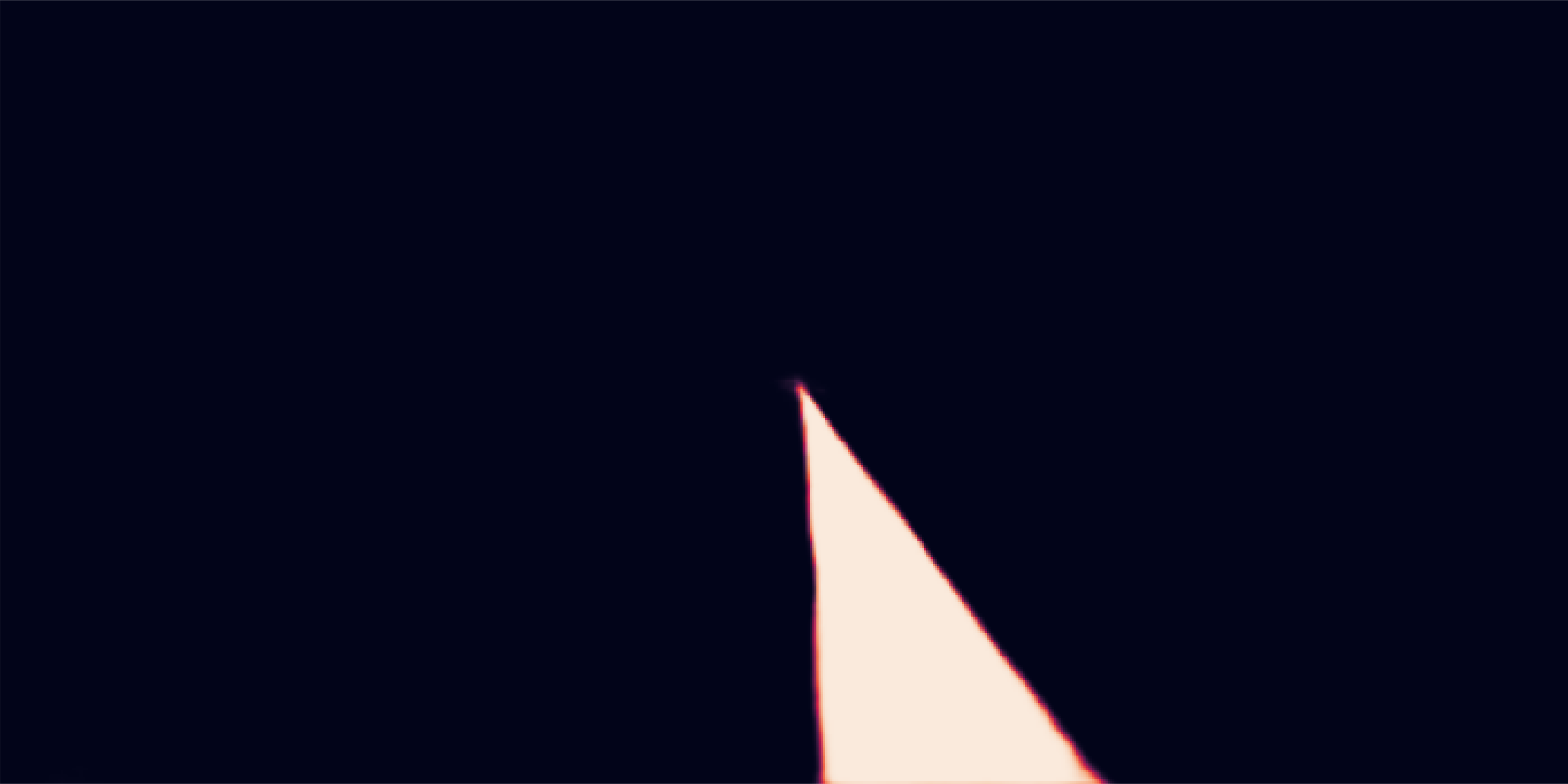} }}%
    \hfill
    \subfloat[\centering Perturbed]{{\includegraphics[width=.23\textwidth]{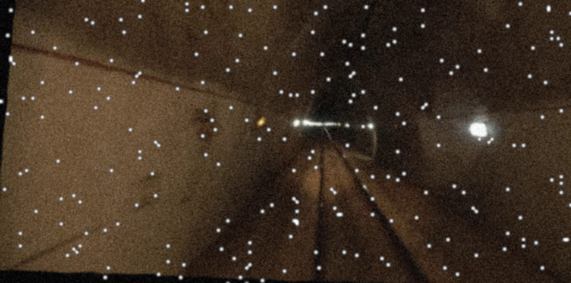} }}%
    \hfill
    \subfloat[\centering IoU = 0]{{\includegraphics[width=.23\textwidth]{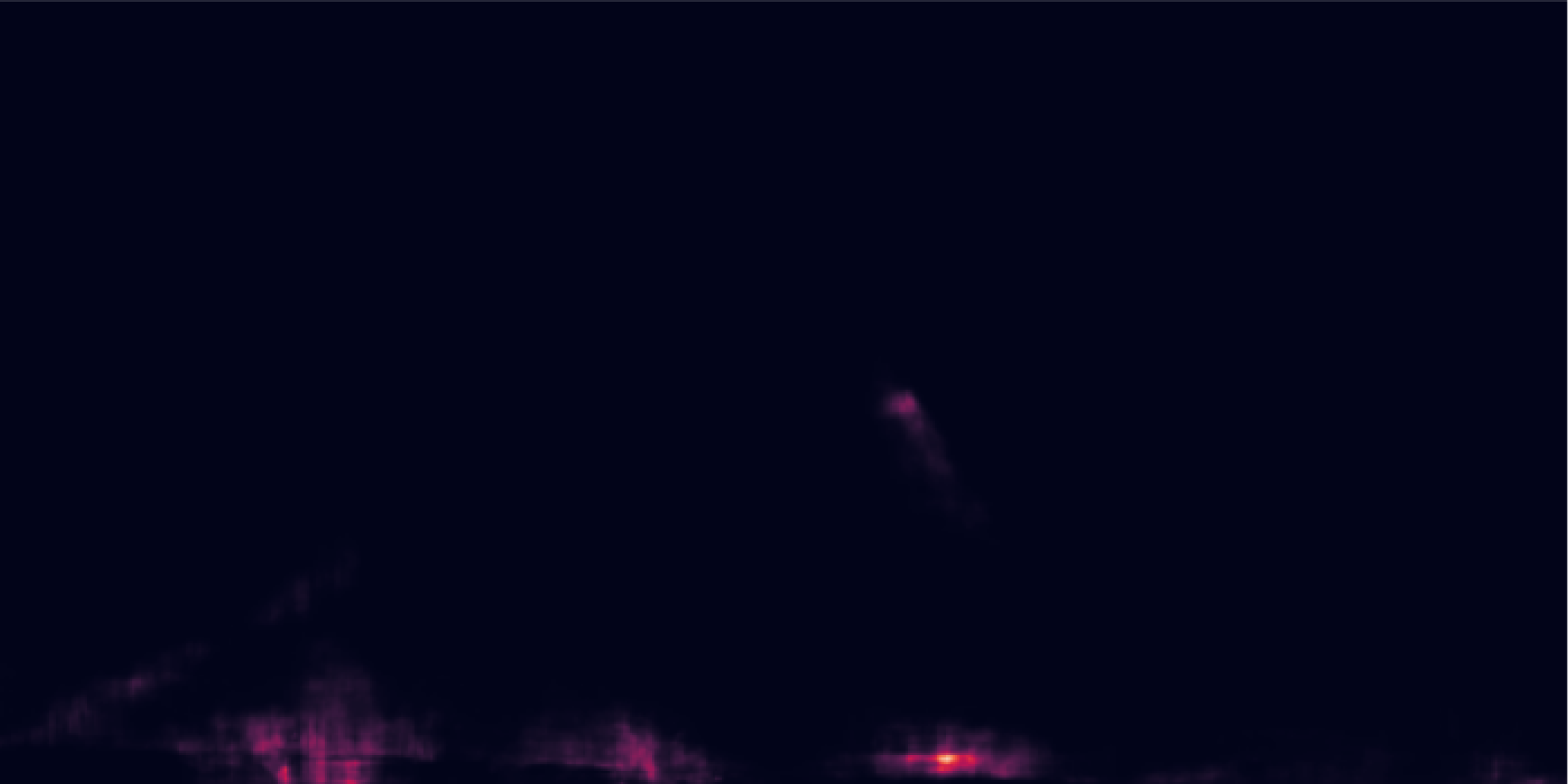} }}%
    \caption{\small Falsifying a given U-net model, which was trained to segment railway tracks. 
    (a, e) show original (unperturbed) images, (b, f) being the respective model output which is the predicted position of railway tracks.  
    Predictions with probability $\le 0.5$ are shown in red and do not contribute to the IoU (see eq.\ref{eq:iou}).
    The approach in Sect.\ref{sec:our_approach} created the perturbed images (c, g) where the model output (d, h) 
    shows maximal deterioration.}%
    \label{fig:example}%
\end{figure}

The approach is evaluated on an image segmentation task in the railway domain. 
Specifically, we are interested in detecting railway tracks in images taken from the ego-perspective of a train as shown in Fig.\ref{fig:example}, where
a tested U-net model achieves a high Intersection-over-Union (IoU) accuracy on the original images but
fails on counterexamples found with the approach described in Sect.\ref{sec:our_approach}.

\section{Related work}
\paragraph{Gradient-based adversarial attacks and $\ell^p$-robustness}
\cite{szegedy2013intriguing} made the observation that the performance of neural networks is prone to small changes in the feature space that are even unperceivable to human eyes. \cite{goodfellow2014explaining} designed a simple fast gradient sign method (FGSM) attack, making use of only the gradient sign, to fool neural networks efficiently. \cite{mkadry2017towards} and \cite{carlini2017towards} proposed iterative approaches known as projected gradient descent (PGD) and Carlini-Wagner (CW) attacks, respectively, to maximize the model's error while keeping the degree of distortion as small as possible. These attacks have been instantly applied to train neural networks (\cite{shafahi2019adversarial, wong2020fast}) to resist potential attacks, by performing attacks during each training epoch. This class of attacks operates exclusively in the original feature space, e.g., on pixel values in case of image data, which can be formulated as an $\ell^p$ space. To this end, they are also referred to as $\ell^p$ attack and the model's capability to resist such attack is known as $\ell^p$-robustness. 

\paragraph{Robustness against natural perturbations}
Instead of the \emph{raw features} that are perturbed by the adversarial attacks, a second class of data distortion focus on altering the \emph{meta features}, which is high-level information of the data. For instance, the brightness and contrast may be meta features of an image. The amplitude of all frequencies may be meta features of time series data. Since these perturbations are realistic in that they can be indeed observed in nature, they are also known as natural perturbations. \cite{hendrycks2019robustness} summarized a set of benchmarking perturbation functions, which in turn serves as additional data augmentation methods \cite{rebuffi2021data}. Due to the fact that such natural perturbations form an open set and that there is already a very large variety thereof, it is very expensive to directly use them as data augmentation.  
\cite{hendrycks2019augmix} and \cite{tang2020selfnorm} developed efficient methods to combine different perturbations into one augmentation to speed up the augmented training. 

DeepTest \cite{DeepTest2018} is close to our approach in the sense that natural perturbations are combined. However, it uses a random search over pairs of only two perturbations to maximize the neuron coverage of the tested DNN. Neuron coverage has been criticized as a suitable measure \cite{Harel2020}. Instead, we directly maximize the model error using gradient-free optimization on chains of up to six perturbations.

\section{Gradient-free optimization of perturbations}\label{sec:our_approach}
\begin{figure}[tbh]%
    \centering
    \includegraphics[width=.7\textwidth]{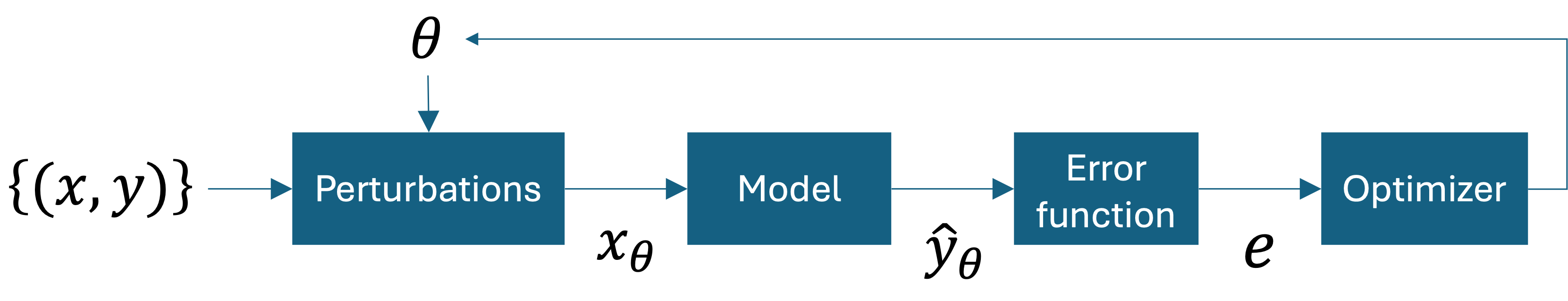}%
    \caption{Optimizing parameter vector $\bf \theta$ to maximize model error $e$}%
    \label{fig:scheme}%
\end{figure}

The objective is to find a sequence of perturbations and their parameters (e.g., mean and standard deviation for Gaussian noise) that maximizes the model error on a set of images. We assume that a set $P$ of perturbations is given ($|P|=12$ in this paper). The optimizer shall draw a subset $K (|K|=6)$ of them in a specific order without replacement, and optimize their parameters. The selection of $K$ with the respective order and all individual parameters are encoded in a total parameter vector $\theta$, which is optimized by a gradient-free optimizer as shown in Fig.\ref{fig:scheme}, where $x$ denotes an input image and $y$ the corresponding label. For the considered task of image segmentation, $y$ is a binary image mask, indicating whether a pixel belongs to the railway tracks. A chain of perturbations $\pi(x,y|\theta)$ yields a perturbed image $x_\theta$ and its
corresponding label $y_\theta$ with $y_\theta \neq y$ only for geometric perturbations (e.g., affine transformations). 
The model predictions on the original image and the perturbed one are $\hat{y}$ and $\hat{y}_\theta$, respectively. The error $e$ maximized by the optimizer is the average deterioration of the IoU,
\begin{equation}\label{eq:error}
    e = N^{-1} \sum_{(x,y)} IoU(\hat{y}, y) - IoU(\hat{y}_\theta, y)
\end{equation}
with $N$ as the number of considered images. The IoU is calculated and averaged over three threshold values, 0.5, 0.9, and 0.99,
\begin{equation}\label{eq:iou}
    IoU(\cdot, y) = \sum_{\tau \in \{0.5, 0.9, 0.99\}} \frac{|I(\cdot > \tau) \land y|}{|I(\cdot >\tau) \lor y|} / 3,
\end{equation}
where $I$ denotes a boolean indicator function, to alleviate potential issues with miscalibrated probabilities of the model predictions.

\subsection{Bounding the perturbation strength}
To avoid trivial counterexamples, like simply turning the contrast to zero, the strength of the perturbations must be bounded. The bounds are defined upfront by a grid-search performed separately for each perturbation parameter: for each parameter value on a grid, the perturbation is applied to the whole dataset, and the IoU of the model on this perturbed dataset is compared to the IoU on the original, unperturbed dataset. The value that yields an average IoU deterioration close to a 1\% is chosen as the maximum limit of that perturbation parameter. That is, we allow every individual perturbation parameter to deteriorate the IoU on average across all images maximally by about 1\%. 

\subsection{Clustering the input images} \label{sec:cluster}
Basically, the approach can be applied to a single image, but the found perturbations are expected to be more meaningful if they are based on multiple images. The optimization results depend on the kind of images. For example, low-light night images are more susceptible to perturbations reducing the contrast. Thus, we run the optimization on clusters of "similar" images, where similarity is defined by firstly extracting features from each image using a pre-trained neural network like EfficientNetB4 \cite{EfficientNet}. This yields a feature vector per image (with length 1700 for EfficientNetB4), whose dimension is further reduced to 10 by UMAP \cite{McInnes2018} with a cosine metric before passing it to a common clustering method like kmeans, hierarchical clustering, or DBSCAN. We prefer kmeans as hierarchical clustering created rather imbalanced clusters and the density-based DBSCAN dropped many images as outliers in low-density regions.

\section{Results} \label{sec:results}
The approach has been tested on the public RailSem19 \cite{zendel_railsem19:_2019} dataset containing 8500 RGB-pictures with dimension 1920 x 1080 taken from the ego-perspective of a train in different environments. The system under test is a Pytorch U-net image segmentation model, which had been trained on that dataset.

While there are public perturbation tools available (e.g., Albumentations, Imgaug), we use an internal tool, which directly supports the run-length encoded label format in RailSem19. 

There are no particular restrictions on the type of perturbations; we just exclude those with too strong effects like histogram equalization, which would make it too easy for the optimizer. Most perturbations have a parameter vector controlling their behavior, e.g., a "rain" perturbation has a 6-dimensional vector with {\small (opaqueness, size, density, blur, angle, speed)} of floats and integers. Perturbations with strong random effects not controlled by a parameter should be avoided as this can hinder the systematic exploration of the parameter space. E.g., "zoom" originally selected a rectangle position purely randomly; we changed this to include the coordinates in its parameter vector to make them accessible to the optimizer.

Images are assigned to 30 clusters by kmeans as described in Sect.\ref{sec:cluster}.  The gradient-free optimizer uses the two-points differential evolution in \cite{nevergrad} with a budget of 5000 iterations on each cluster. With this budget, the verification of a cluster of, e.g., 300 images takes about 20 hours on a GPU processing ca.\ 20 images per second.
\smallskip

\begin{figure}[ht]
    \centering
    \subfloat[\centering Clusters]{%
    \adjustbox{valign=c}{%
    \begin{minipage}{0.49\textwidth}%
    \scriptsize%
    \begingroup%
    \setlength{\tabcolsep}{2pt}
    \renewcommand{\arraystretch}{0.95}
    \begin{tabular}{c|r|l|c }%
        \hline%
        \textbf{ID} & \textbf{Size} & \textbf{Description} & \textbf{Deter.} \\ \hline \hline%
        0       & 631   & City streets                     & \cellcolor{brown!40} 0.42 \\ \hline%
        1       & 431   & Rails on gravel                  & \cellcolor{brown!00} 0.10 \\ \hline%
        2       & 249   & Uncertain                        & \cellcolor{brown!00} 0.10 \\ \hline%
        3       & 313   & Suburbs                          & \cellcolor{brown!10} 0.14 \\ \hline%
        4       & 386   & Beams, cloudy *                  & \cellcolor{brown!30} 0.33 \\ \hline%
        5       & 383   & Bridges                          & \cellcolor{brown!20} 0.20 \\ \hline%
        6       & 419   & Snow                             & \cellcolor{brown!20} 0.22 \\ \hline%
        7       & 446   & Beams, clear skies               & \cellcolor{brown!10} 0.15 \\ \hline%
        8       & 130   & Trees                            & \cellcolor{brown!10} 0.12 \\ \hline%
        9       & 97    & Snow with wiper                  & \cellcolor{brown!40} 0.44 \\ \hline%
        10      & 309   & Night / tunnels *                & \cellcolor{brown!60} 0.65 \\ \hline%
        11      & 388   & Yellow fields                    & \cellcolor{brown!00} 0.08 \\ \hline%
        12      & 92    & Brown soil w/ wiper *            & \cellcolor{brown!40} 0.46 \\ \hline%
        13      & 145   & Train / railcar                  & \cellcolor{brown!10} 0.15 \\ \hline%
        14      & 209   & Trees                            & \cellcolor{brown!10} 0.14 \\ \hline%
        15      & 507   & Uncertain                        & \cellcolor{brown!10} 0.16 \\ \hline%
        16      & 397   & Grasses                          & \cellcolor{brown!10} 0.11 \\ \hline%
        17      & 104   & Grass with wiper                 & \cellcolor{brown!00} 0.08 \\ \hline%
        18      & 210   & Window w/ wiper *                & \cellcolor{brown!40} 0.42 \\ \hline%
        19      & 347   & Urban / cars                     & \cellcolor{brown!30} 0.33 \\ \hline%
        20      & 444   & Rails on gravel w/ trees         & \cellcolor{brown!00} 0.08 \\ \hline%
        21      & 140   & Grass or snows w/ wiper          & \cellcolor{brown!20} 0.26 \\ \hline%
        22      & 195   & Cloudy evenings *                & \cellcolor{brown!30} 0.34 \\ \hline%
        23      & 257   & Train / railcar present          & \cellcolor{brown!10} 0.17 \\ \hline%
        24      & 83    & Suburbs w/ wiper                 & \cellcolor{brown!10} 0.16 \\ \hline%
        25      & 443   & Suburbs                          & \cellcolor{brown!10} 0.14 \\ \hline%
        26      & 128   & Traffic lights, cloudy           & \cellcolor{brown!20} 0.22 \\ \hline%
        27      & 118   & Arched bridge                    & \cellcolor{brown!10} 0.17 \\ \hline%
        28      & 190   & Mountains                        & \cellcolor{brown!10} 0.18 \\ \hline%
        29      & 309   & Square bridge / buildings        & \cellcolor{brown!10} 0.10 \\ \hline%
    \end{tabular}%
    \endgroup%
    \end{minipage}%
    \label{fig:table_clusters}%
    }}%
    \hfill%
    \begin{minipage}{0.5\textwidth}%
    \subfloat[\centering Perturbations]{%
    \adjustbox{valign=c}{%
        \includegraphics[width=\textwidth]{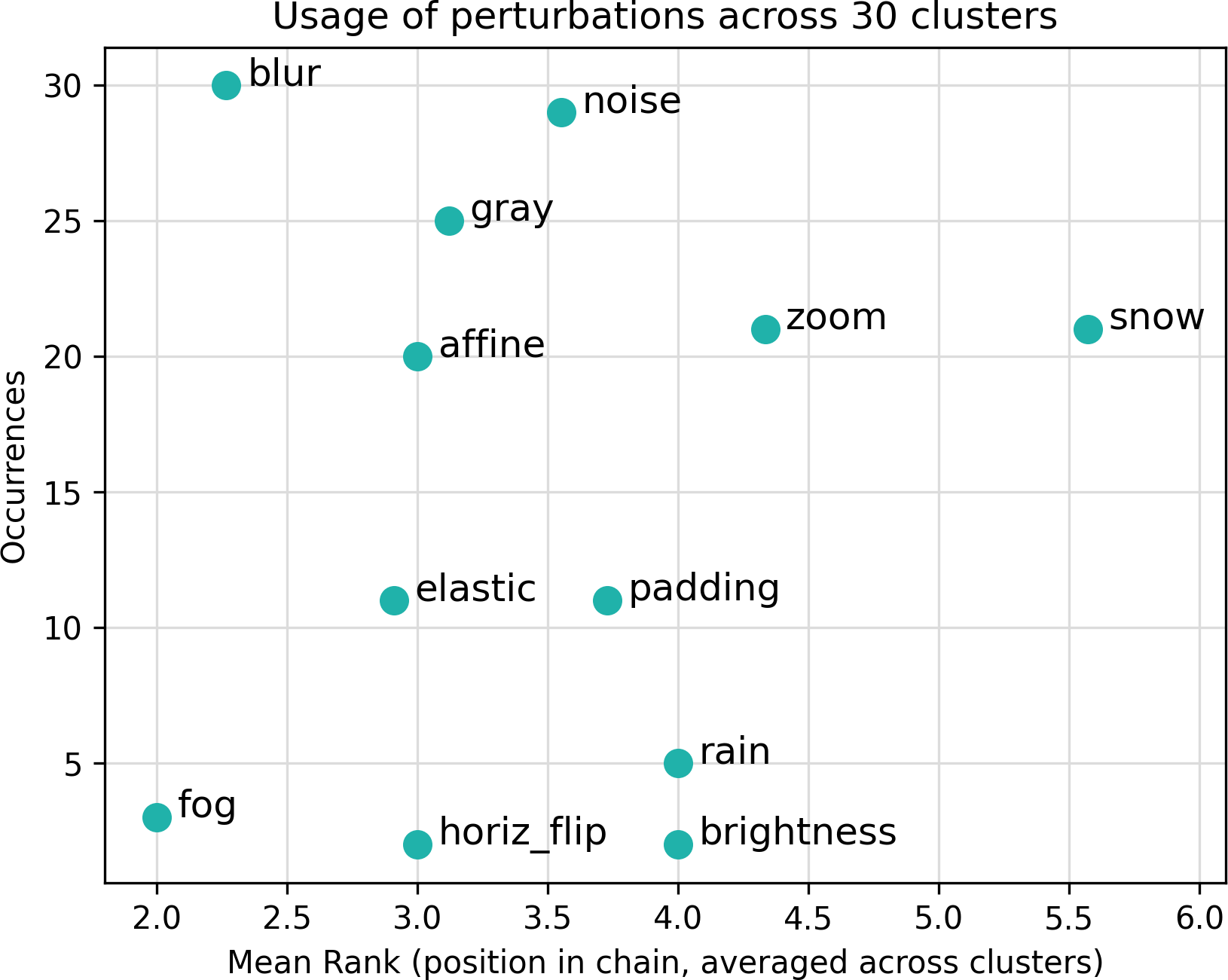}\label{fig:perturbation}%
    }}\\[5mm]
    \caption{a) Mean IoU-deterioration per cluster {\small (descriptions with * indicate brightness perturbation being disabled to avoid trivial counterexamples)}. Examples (a) and (e) in Fig.\ref{fig:example} belong to clusters 9 and 10, respectively. b)~Usage of perturbations across clusters.}%
    \end{minipage}
\end{figure}

The mean IoU-deterioration per cluster in Fig.\ref{fig:table_clusters} shows that the model is less robust on dark images, images with wipers (e.g., Fig.\ref{fig:ex1_orig}), and complex urban environments. Note that on every cluster a separate optimization is performed, yielding a cluster-specific chain of perturbations. E.g., in cluster 9, where Fig.\ref{fig:ex1_orig} belongs to, the chain was found to be: {\small ("blur", "noise", "affine", "zoom", "rain", "padding")}. Fig.\ref{fig:perturbation} shows that "blur" is chosen in every cluster and as one of the first perturbations in the chain. In contrast, "snow" is added lately (probably to avoid diminishing its effect by later perturbations). "Fog" only occurs in three clusters but as one of the first perturbations. More details can be found in \cite{oswan2023}. 

\section{Conclusion}
Gradient-free optimization can find perturbation chains that successfully falsify a given model, going beyond existing grid-based or combinatorial testing. The approach can support the verification by identifying weaknesses of the model regarding specific combinations of perturbations on specific types of images. The insights can also be used to refine the training process of the model, e.g., by adding perturbed training examples from the clusters with high IoU deterioration.

\paragraph{Acknowledgement} We thank Benjamin Hartmann from Siemens Mobility for helpful discussions and for providing the U-net model tested in this approach. This research has received funding from the German Federal Ministry for Economic Affairs and Climate Action under grant agreement 19I21039A (safe.trAIn).
\bigskip

\begin{footnotesize}

\end{footnotesize}

\end{document}